\documentclass[conference]{IEEEtran}
\IEEEoverridecommandlockouts
\usepackage{amsmath,amssymb,amsfonts}
\usepackage{algorithmic}
\usepackage{graphicx}
\usepackage{textcomp}
\usepackage{xcolor}

\usepackage{amsmath,amssymb,amsfonts}
\usepackage{algorithmic}
\usepackage{graphicx}
\usepackage{textcomp}
\usepackage{xcolor}
\usepackage{multirow}

\usepackage{subfig}
\usepackage{dblfloatfix}    

\usepackage{graphicx}

\usepackage{hyperref} 

\usepackage[linesnumbered, ruled]{algorithm2e}

\usepackage[labelsep=period]{caption} 

\usepackage[bottom]{footmisc} 

\DeclareMathOperator{\EX}{\mathbb{E}}

\usepackage[
style=ieee,
backend=biber,
sorting=none
]{biblatex}

\def\BibTeX{{\rm B\kern-.05em{\sc i\kern-.025em b}\kern-.08em
    T\kern-.1667em\lower.7ex\hbox{E}\kern-.125emX}}
\begin{document}

\title{Decaying Clipping Range in Proximal Policy Optimization}

\author{\IEEEauthorblockN{Mónika Farsang}
\IEEEauthorblockA{\textit{Department of Automation and Applied Informatics} \\
\textit{Budapest University of Technology and Economics}\\
Budapest, Hungary \\
monika.farsang@edu.bme.hu}
\and
\IEEEauthorblockN{Dr. Luca Szegletes}
\IEEEauthorblockA{\textit{Department of Automation and Applied Informatics} \\
\textit{Budapest University of Technology and Economics}\\
Budapest, Hungary \\
luca.szegletes@aut.bme.hu}
}

\maketitle

\begin{abstract}
Proximal Policy Optimization (PPO) is among the most widely used algorithms in reinforcement learning, which achieves state-of-the-art performance in many challenging problems. The keys to its success are the reliable policy updates through the clipping mechanism and the multiple epochs of minibatch updates. The aim of this research is to give new simple but effective alternatives to the former. For this, we propose linearly and exponentially decaying clipping range approaches throughout the training. With these, we would like to provide higher exploration at the beginning and stronger restrictions at the end of the learning phase. We investigate their performance in several classical control and locomotive robotic environments. During the analysis, we found that they influence the achieved rewards and are effective alternatives to the constant clipping method in many reinforcement learning tasks.
\end{abstract}

\begin{IEEEkeywords}
Clipping range, Proximal Policy Optimization, Reinforcement learning, Robotic control
\end{IEEEkeywords}

\section{Introduction}

In reinforcement learning, the goal of the agent is to find the optimal policy to maximize the obtained rewards by interacting with the environment. This field gives the agents the ability to learn and adapt to many robotic and control problems \cite{pmlr-v48-duan16} based on exploration. Frequently used algorithm types are the policy gradient methods.

Policy gradient methods compute an estimator of the gradient of the expected return collected by sample trajectories. At the same time, they often have an undesirable behavior with large policy updates. To guarantee monotonic improvement, Trust Region Policy Optimization (TRPO) \cite{pmlr-v37-schulman15} was introduced as a modification of the classical natural policy gradient algorithm \cite{Kakade2001}. By applying  Kullback–Leibler divergence rather than a fixed penalty coefficient, it gives a stable and reliable policy gradient approach. However, it uses a second-order optimization technique, which leads to difficult scaling and high computation complexity.

To address these limitations of TRPO, Schulman \textit{et al.} \cite{schulman2017proximal} proposed the robust PPO algorithm, which restricts the changes to the policy that are far away from the old one by clipping the probability ratio. This method has two main parts, which are the data sampling and the multiple epochs of minibatch updates with optimization of the surrogate objective function. During the optimization, stochastic gradient ascent is used. Many advantages come from the approach of PPO compared to TRPO: it is simpler to implement, scalable, has better sample complexity and better performance.

It was also demonstrated in \cite{schulman2017proximal} that PPO outperforms several other online policy gradient methods in many continuous control environments. They compared the results of the vanilla policy gradient with adaptive stepsize, the adaptive TRPO and the Cross-Entropy Method (CEM) \cite{szita2006}. They also examined the Advantage Actor Critic (A2C) \cite{mnih2016}, which is the synchronous variant of the Asynchronous Advantage Actor Critic (A3C) algorithm, and its improved variant with trust region \cite{wang2017sample}.

In recent years, different improvements in the PPO algorithm were suggested as well. The significance of state-action pairs is considered by \cite{chen2018adaptive}, where the adaptive control is handled by a new hyperparameter. Wang \textit{et al.} modified PPO with a trust region guided criterion \cite{wang2019trust}, which requires additional computation. 

To maintain the simple approach of PPO without introducing any other hyperparameter or adding extra computational load to the algorithm, we propose linearly and exponentially decaying clipping ranges. PPO with constant clipping range proved to be possibly leading to insufficient exploration \cite{wang2019trust}. With our approaches, we would like to give sufficient exploration at the beginning of the training and stronger restrictions in the policy updates at the end.

This paper is divided into five sections. This section was a brief overview of the policy gradient methods and their challenges. The next section describes the essential background of this study. Our design and implementation of our methodology with the classical control and robotic environments are outlined in the third section. The fourth section analyses the results of our new approach. Our conclusions are drawn in the final section.

\section{Background}
Reinforcement learning problems can be described as Markov Decision Processes, where the policy $\pi$ maps each state $s_t$ from the state space $S$ to action $a_t$ from the action space $A$ at timestep $t$. 
In policy gradient methods, two important concepts are the probability ratio \eqref{eq_ratio} and the advantage function \eqref{eq_advantage}. The former measures the changed probability of the chosen action $a_t$ in state $s_t$ under the new and the old policy. The policy is modeled with a parameterized function $\pi_{\theta}$.
\begin{equation}
r_t(\theta)=\frac{\pi_{\theta}(a_t \mid s_t)}{\pi_{{\theta}_{old}}(a_t \mid s_t)}\label{eq_ratio}
\end{equation}
 
The advantage function  gives the relative advantage of action $a$ in state $s$ with the action-value $Q^{\pi}(s,a)$ compered to selecting any action in state $s$, formulated by the state-value function $V^{\pi}(s)$.
\begin{equation}
A^{\pi}(s,a)=Q^{\pi}(s,a)-V^{\pi}(s)\label{eq_advantage}
\end{equation}

The multiplication of the two before-mentioned terms formulates \eqref{eq_objtrpo}, the objective function of TRPO algorithm \cite{pmlr-v37-schulman15}, where $\hat{A_t}$ is an estimator of the advantage function at timestep $t$ and $\hat{\EX}_t$ is the empirical average over a finite batch of samples.

\begin{equation}
L^{TRPO}(\theta)=\hat{\EX}_t\left[r_t(\theta)\hat{A_t}\right]
\label{eq_objtrpo}
\end{equation}

In PPO \cite{schulman2017proximal}, the objective function is basically the same as \eqref{eq_objtrpo} if the probability ratio is close to $1$, which means, that the new and the old policy are close to each other. However, this method uses a clipping approach to those situations, where the new policy is far away from the old one. This clipping parameter is denoted by $\epsilon$ and it prevents large updates in the policy, which has a value usually around 0.2. The minimum term is used to give a pessimistic lower bound to the policy surrogate. This simple yet powerful idea prevents large policy updates during optimization.

\begin{equation}
L^{PPO}(\theta)=\hat{\EX}_t\left[\text{min}(r_t(\theta)\hat{A_t}, \text{clip}(r_t(\theta),1-\epsilon, 1+\epsilon)\hat{A_t})\right]\label{eq_objppo}
\end{equation}

Loss function \eqref{eq_totallossppo} is defined by the combination of three terms \cite{schulman2017proximal}: the previously described surrogate function \eqref{eq_objppo}, the value function $L^{VF}$ and the entropy bonus $S$, as proposed by earlier research \cite{Williams1992, mnih2016}, where $c_1$ and $c_2$ are two coefficients.

\begin{equation}
L_t(\theta)=\hat{\EX}_t\left[L^{PPO}_t(\theta)-c_1L^{VF}_t+c_2S[\pi_\theta](s_t)\right]\label{eq_totallossppo}
\end{equation}

The optimization is based on this term with respect to $\theta$. As a new approach proposed by Schulman \textit{et al.} \cite{schulman2017proximal}, there are multiple epochs of minibatch updates, compared to TRPO, where there is only one gradient update per data sample.

\section{Methods}
This section begins by describing the six classical control and robotic RL environments that we used for training. The next subsections outline the design and implementation of the PPO with the new clipping approaches.

\subsection{Environments}\label{Env}
\begin{figure}[!t]
\captionsetup{justification=centering}
  \begin{minipage}{.3\linewidth}
\subfloat[\label{fig:openaigyma}]{\includegraphics[height=\linewidth, width=\linewidth]{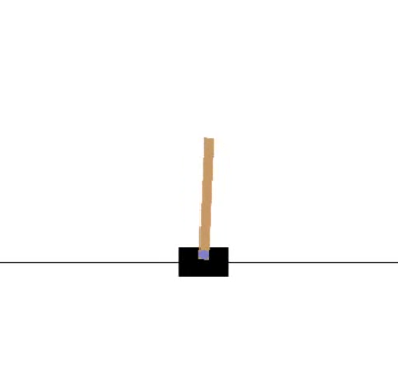}}
  \end{minipage}%
  \hfill
  \begin{minipage}{.3\linewidth}
    \subfloat[\label{fig:openaigymb}]{\includegraphics[height=\linewidth, width=\linewidth]{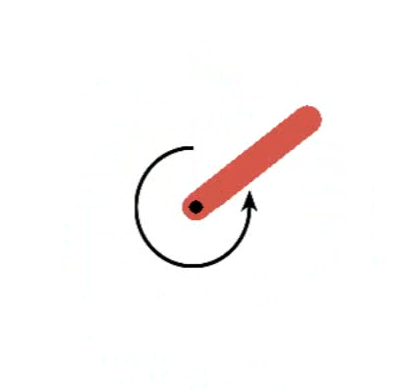}}
  \end{minipage}%
  \hfill
    \begin{minipage}{.3\linewidth}
\subfloat[\label{fig:openaigymc}]{\includegraphics[height=\linewidth, width=\linewidth]{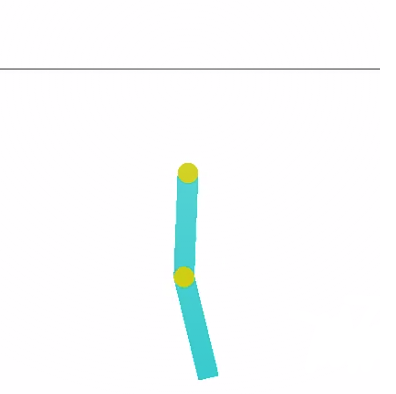}}
  \end{minipage}%
  \hfill
    \begin{minipage}{.3\linewidth}
\subfloat[\label{fig:pybulletd}]{\includegraphics[height=\linewidth, width=\linewidth]{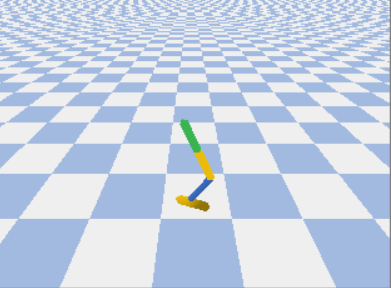}}
  \end{minipage}%
  \hfill
  \begin{minipage}{.3\linewidth}
    \subfloat[\label{fig:pybullete}]{\includegraphics[height=\linewidth, width=\linewidth]{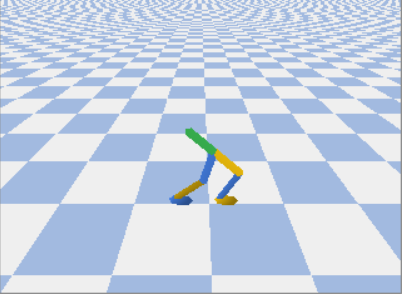}}
  \end{minipage}%
  \hfill
    \begin{minipage}{.3\linewidth}
\subfloat[\label{fig:pybulletf}]{\includegraphics[height=\linewidth, width=\linewidth]{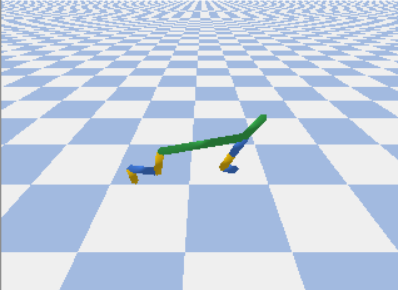}}
  \end{minipage}%
  \caption{OpenAI Gym classical control and PyBullet robotic environments. (a) Cart-Pole. (b) Pendulum. (c) Acrobot. (d) Hopper. (e) Walker. (f) Half-Cheetah \cite{brockman2016openai, coumans2017pybullet}}
  \label{fig:openaigympybullet}
\end{figure}

We used the OpenAI Gym \cite{brockman2016openai} for training agents in classical control tasks. The three selected exercises are the Cart-Pole, the inverted Pendulum and the Acrobot control problems\footnote{CartPole-v1, Pendulum-v0 and Acrobot-v1}. They can the seen in Fig. \ref{fig:openaigympybullet}\subref{fig:openaigyma}, \ref{fig:openaigympybullet}\subref{fig:openaigymb} and \ref{fig:openaigympybullet}\subref{fig:openaigymc}.

The Cart-Pole problem was introduced by Sutton and Barto \cite{Sutton1998}. The goal is to balance the pole, which is attached to a cart, and prevent it from falling. The position and the velocity of the cart and the angle and velocity at the tip of the pole are observed. The possible actions are pushing the cart to the left or to the right. 

Fig. \ref{fig:openaigympybullet}\subref{fig:openaigymb} shows the Pendulum task, where the aim is to try to keep it in the standing position by applying torque to its joint. The observation space consists of the pendulum angle and the angular velocity.

The Acrobot, which was first described by Sutton \cite{Sutton1996}, has two links and joints. This version of the environment uses Runge-Kutta integration for better accuracy \cite{geramifard2015rlpy}. The joint between the links hangs downwards at the initial state, which has to be controlled in order to bring the lower link up to the given height. The environment is similar to the previous one, the two rotational joint angles and the joint angular velocities are observed and the action is applying torque on the controllable joint.

For more complex problems, we used PyBullet \cite{coumans2017pybullet}, which provides several robotic environments. We selected the Hopper, the Walker and the Half-Cheetah locomotion tasks\footnote{HopperBulletEnv, Walker2DBulletEnv and HalfCheetahBulletEnv, all with version "-v0"}, which are presented in Fig. \ref{fig:openaigympybullet}\subref{fig:pybulletd}, \ref{fig:openaigympybullet}\subref{fig:pybullete} and \ref{fig:openaigympybullet}\subref{fig:pybulletf}. With these, we would like to investigate the performance of the different clipping strategies on high dimensional state and action spaces with increasing difficulty.

Hopper, the one-legged robot with the task of hopping forward, was analyzed in \cite{murthyraibert1984}, \cite{pmlr-v28-levine13}. This model of the hopping robot is based on the work by Erez, Tassa and Todorov \cite{Erez2011} with 15-dimensional observation space and with torque control on the foot, the leg and the thigh.

Many studies have been published on the Walker bipedal robot \cite{pmlr-v37-schulman15}, \cite{raibert1991, pmlr-v28-levine13, Erez2011}. It provides a more advanced environment with torque control on both legs and observation space with 17 dimensions. As the name suggests, the goal is to achieve forward walking motion.

The two-dimensional robot has to learn how to run by controlling the torque on the two feet, shins and thighs in the Half-Cheetah environment. This problem was studied in \cite{wawrzynski2007}, \cite{Heess2015}. The dimension of the observation space is 26, which is significantly larger than in the previously described classical control and even the two other locomotive tasks. 

\subsection{Algorithm}\label{Alg}

Our goal was to maintain the simple design of the PPO algorithm without modifying the objective function and without introducing any extra hyperparameters. Instead of having constant clipping value throughout the learning, we decay this parameter linearly or exponentially. Equation \eqref{eq_linear} shows the linear approach, where $T$ stands for the maximum timestep and $t$ for the current timestep of the training. The initial clipping value is denoted by $\epsilon_0$. 
The selected exponential method \eqref{eq_exp} uses slower decay with $\alpha=0.99$. During the analysis, we compare these two alternatives to the classical PPO method.

\begin{equation}
\epsilon_t^{lin}=\frac{T-t}{T}\epsilon_0\label{eq_linear}
\end{equation}

\begin{equation}
\epsilon_t^{exp}=\alpha^{100\frac{t}{T}}\epsilon_0\label{eq_exp}
\end{equation}

The modified version of the PPO algorithm is shown in Algorithm \ref{algo_ppo}. The new element is placed in line 6, the other parts originate from \cite{schulman2017proximal}.

\begin{algorithm}
\For{iteration = 1, 2, ... }{
\For{actor = 1, 2, ..., N}{
 Run policy $\pi_{\theta_{old}}$ in environment for $T$ timesteps\\
 Compute advantage estimates $\hat{A}_1 \dots \hat{A}_T$\
}
Decay the clipping range linearly \eqref{eq_linear} or exponentially \eqref{eq_exp}\\
Optimize surrogate $L$ wrt $\theta$, with $K$ epochs and minibatch size $M \leq NT$\\
$\theta_{old} \xleftarrow[]{} \theta_{new}$\
}
\caption{PPO algorithm with clipping decay}\label{algo_ppo}
\end{algorithm}

\subsection{Implementation}\label{Impl}
Stable Baselines3 (SB3) \cite{stable-baselines3} provides reliable implementations of reinforcement learning algorithms in PyTorch to give good baselines and make it easier to refine new ideas. This code is based on exactly the algorithm described by Schulman \textit{et al.} \cite{schulman2017proximal}. Because of these characteristics, we found it a perfect choice and implemented the alternative clipping approaches to the PPO algorithm.  Source code is available at \url{https://github.com/MoniFarsang/ppo-clipping-approaches}.

\section{Results}

\begin{figure}[!b]
\centering
\captionsetup{justification=centering}
\includegraphics[width=.7\linewidth]{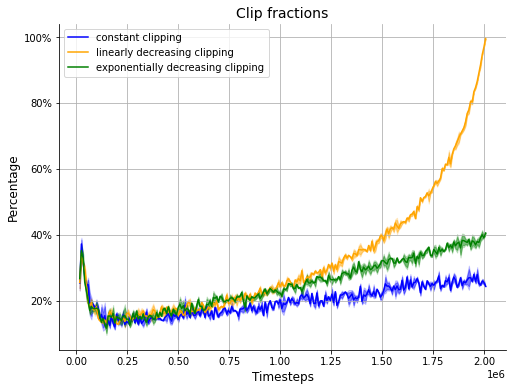}
\caption{Percentage of clipping the loss function in policy updates throughout the training}
\label{fig:clips}
\end{figure}

First, we analyzed the mechanism of the exponential and linear decaying clipping. We found that they influence strongly the policy updates during training. It is demonstrated by Fig. \ref{fig:clips}, which displays the percentage of existing clips in practice. Besides the constant value, slow exponential decay and linear decay to zero are illustrated. The fractions of the truncated updates are higher due to the poor performance at the beginning of the training. Later, they decrease then slowly start to rise again. The reason behind this comes from two aspects: the policy updates more often exceed the limit and the clipping range decays by the algorithm. The constant clipping value influences only the first aspect, which gives a good frame of reference. Accordingly, the differences between the two other curves show the effect of the decreasing limits. 

\begin{figure*}[!tb]
\captionsetup{justification=centering}
  \begin{minipage}{.29\linewidth}
\subfloat[\label{fig:openairesa}]{\includegraphics[height=\linewidth, width=\linewidth]{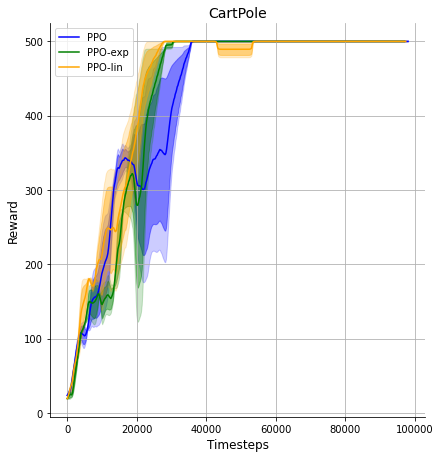}}
  \end{minipage}%
  \hfill
  \begin{minipage}{.29\linewidth}
    \subfloat[\label{fig:openairesb}]{\includegraphics[height=\linewidth, width=\linewidth]{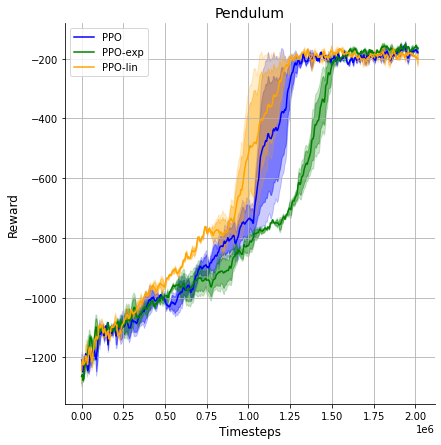}}
  \end{minipage}%
  \hfill
    \begin{minipage}{.29\linewidth}
\subfloat[\label{fig:openairesc}]{\includegraphics[height=\linewidth, width=\linewidth]{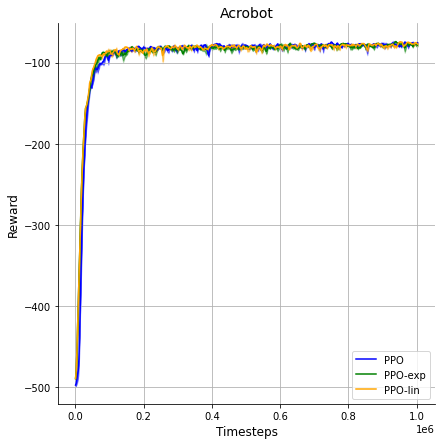}}
  \end{minipage}%
  \hfill
    \begin{minipage}{.29\linewidth}
\subfloat[\label{fig:pybulletresd}]{\includegraphics[height=\linewidth, width=\linewidth]{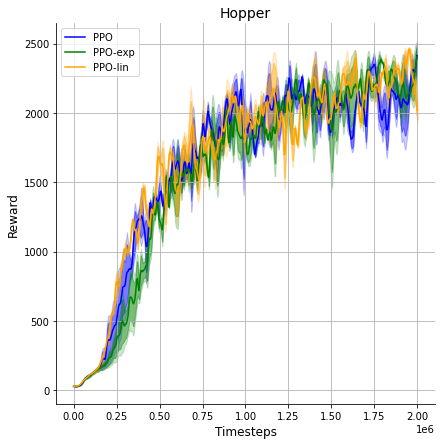}}
  \end{minipage}%
  \hfill
  \begin{minipage}{.29\linewidth}
    \subfloat[\label{fig:pybulletrese}]{\includegraphics[height=\linewidth, width=\linewidth]{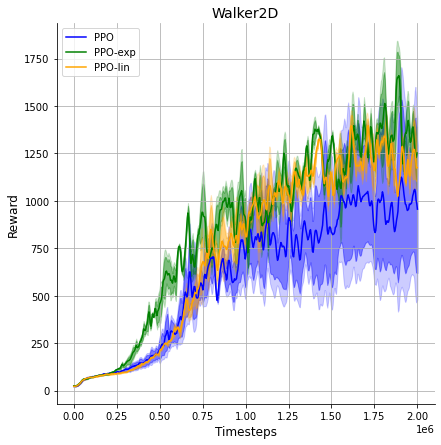}}
  \end{minipage}%
  \hfill
    \begin{minipage}{.29\linewidth}
\subfloat[\label{fig:pybulletresf}]{\includegraphics[height=\linewidth, width=\linewidth]{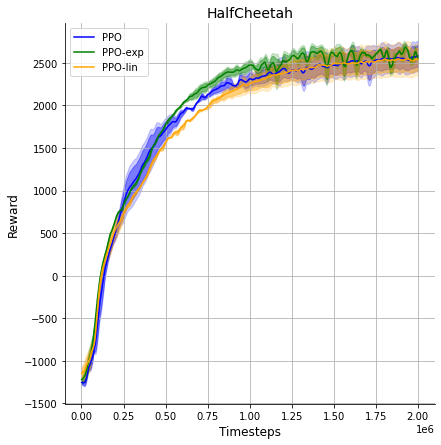}}
  \end{minipage}%
  \caption{Average rewards obtained by different clipping range strategies in OpenAI Gym classical control and in PyBullet robotic environments.}
  \label{fig:openaigym_pybullet_results}
\end{figure*}%

We evaluated the PPO algorithm with three different clipping strategies and our focus was on the average rewards during the training. To compare the different clipping strategies, we use the evaluation metrics proposed by Schulman \textit{et al.}  \cite{schulman2017proximal}. These are the average reward per episode over the entire training period and over the last 100 episodes of training. Meanwhile, the former measures fast learning, the latter analyzes the final performance.

In all our experiments of the OpenAI Gym classical control problems, the proposed linearly and exponentially decaying clipping range give a considerable alternative to the constant value. We present the learning curves in Fig. \ref{fig:openaigym_pybullet_results}\subref{fig:openairesa}, \ref{fig:openaigym_pybullet_results}\subref{fig:openairesb}, \ref{fig:openaigym_pybullet_results}\subref{fig:openairesc}. They were run with 100 thousand, 2 million and 1 million timesteps, respectively. Meanwhile, 8 parallel environments were used in the Cart-Pole and the Pendulum tasks, 16  was in the Acrobot problem. The evaluation metrics are reported in Table \ref{tab_openaigym_pybullet}. We found that the linear decaying clipping range gives the best performance during the whole training in each task. However, if we look at only the last 100 episodes, the exponentially decreasing strategy gives better results in the case of the Pendulum problem and the linear one with the Acrobot. In the latter, the results are close to each other, as can be noted in Fig. \ref{fig:openaigym_pybullet_results}\subref{fig:openairesc}. Similarly, each algorithm got the same results in the Cart-Pole task, where they archived the maximum possible score.

\begin{table*}[!tb]
\captionsetup{justification=centering}
\caption{\\ \textsc{Performance of the clipping strategies in OpenAI Gym and PyBullet environments}}
\begin{center}
\begin{tabular}{|c|c|c|c|c|c|c|c|}
\hline
\textbf{Average} & \textbf{Clipping}&\multicolumn{3}{|c|}{\textbf{OpenAI Gym Environments}}&\multicolumn{3}{|c|}{\textbf{PyBullet Environments}} \\
\cline{3-8} 
\textbf{rewards} & \textbf{decay} & \textbf{Cart-Pole} & \textbf{Pendulum} & \textbf{Acrobot} & \textbf{Hopper} & \textbf{Walker} & \textbf{Half-Cheetah} \\
\hline
\multirow{3}{*}{all training} & constant & 419.85 & -629.01 & -91.42  & 1628.91 &  603.86 & 2536.31  \\
& linear & \textbf{435.10} & \textbf{-579.64} & \textbf{-89.29} & \textbf{1669.25} & 738.83 & 2524.82 \\
& exponential & 422.91 & -698.49 & -90.08 & 1576.96 & \textbf{840.74} & \textbf{2586.07}  \\
\hline
\multirow{3}{*}{last 100 episodes} & constant & \textbf{500.00} & -186.90 & -77.96 & 2108.30 & 984.99 & 1891.02 \\
& linear & \textbf{500.00} & -190.14 & \textbf{-77.95} & \textbf{2230.95} & 1230.14  & 1844.64 \\
& exponential & \textbf{500.00} & \textbf{-170.76} & -78.38 & 2215.78 & \textbf{1356.59} & \textbf{1982.20} \\
\hline
\end{tabular}
\label{tab_openaigym_pybullet}
\end{center}
\end{table*}

The learning curves of the locomotive robotic tasks in the PyBullet environment are presented in Fig. \ref{fig:openaigym_pybullet_results}\subref{fig:pybulletd}, \ref{fig:openaigym_pybullet_results}\subref{fig:pybullete} and  \ref{fig:openaigym_pybullet_results}\subref{fig:pybulletf}. They were trained 2 million timesteps long in 16 parallel environments. Comparable to the classical control problems, the two proposed strategies show excellent performance and cope with larger complexity as well. Table \ref{tab_openaigym_pybullet} compares the numerical results. The training of the hopping robot reported similar performance with each clipping approach and the linear one was only slightly better than the others. However, the Walker and the Half-Cheetah environment give more diverse outcomes. The exponential decay strategy yields great results in these tasks. They not only exceed the other clipping strategies throughout the whole training but affect especially significantly the final performance.

\section{Conclusion}
In conclusion, our work presented a small modification of the clipping range in the classical PPO algorithm. Our goal was to maintain the simple approach of the original method without modifying the objective function or adding extra hyperparameters. We compared three different clipping range strategies, where we found that the proposed linearly and exponentially decaying values give promising alternatives to the constant clipping range. During the analysis, our focus was on classical control and locomotive robotic environments. The results of this study suggest that varying the clipping range can be beneficial, especially for more complex problems to achieve considerable improvement in the final performance.


\printbibliography

\end{document}